%% file: ms.tex
\documentclass[a4paper, 10pt, conference]{ieeeconf}      

\IEEEoverridecommandlockouts                              

\usepackage{graphics} 
\usepackage{epsfig} 
\usepackage{mathptmx} 
\usepackage{times} 
\usepackage{amsmath} 
\usepackage{amssymb}  
\usepackage{pgfplots}
\usepackage{import}
\usepackage{xcolor,colortbl}
\usepackage{array}
\newcolumntype{C}[1]{>{\centering\arraybackslash}p{#1}}
\newcolumntype{G}[1]{>{\columncolor{green}\centering\arraybackslash}p{#1}}
\newcolumntype{R}[1]{>{\columncolor{red}\centering\arraybackslash}p{#1}}

\DeclareMathAlphabet{\mathcal}{OMS}{cmsy}{m}{n}
\renewcommand\vec{\mathbf}

\title{\LARGE \bf
Deep, spatially coherent Inverse Sensor Models with Uncertainty Incorporation using the evidential Framework
}

\author{Daniel Bauer$^{1}$, Lars Kuhnert$^{1}$ and Lutz Eckstein$^{2}$
\thanks{$^{1}$Daniel Bauer and Lars Kuhnert are  with the Ford Werke GmbH, Cologne,
        {\tt\small dbauer31@ford.de}, {\tt\small lkuhnert@ford.de}}%
\thanks{$^{3}$Lutz Eckstein is with the Institute for Automotive Engineering, RWTH Aachen University,
		{\tt\small office@ika.rwth-aachen.de}}%
}

\begin{document}

\maketitle
\thispagestyle{empty}
\pagestyle{empty}

\begin{abstract}

To perform high speed tasks, sensors of autonomous cars have to provide as much information in as few time steps as possible. However, radars, one of the sensor modalities autonomous cars heavily rely on, often only provide sparse, noisy detections. These have to be accumulated over time to reach a high enough confidence about the static parts of the environment. For radars, the state is typically estimated by accumulating inverse detection models (IDMs). We employ the recently proposed evidential convolutional neural networks which, in contrast to IDMs, compute dense, spatially coherent inference of the environment state. Moreover, these networks are able to incorporate sensor noise in a principled way which we further extend to also incorporate model uncertainty. We present experimental results that show This makes it possible to obtain a denser environment perception in fewer time steps.  

\end{abstract}

%
\section{INTRODUCTION}
\begin{figure}
	\begin{center}
		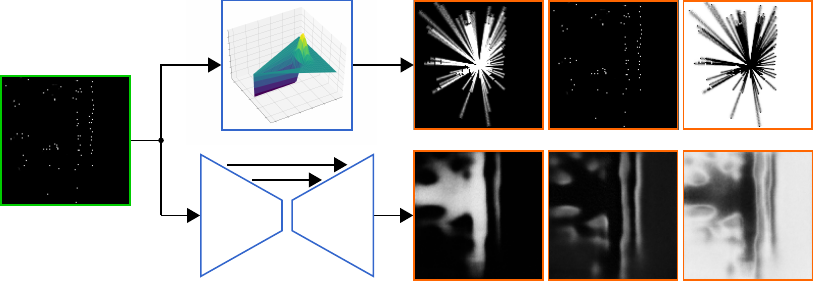
		\caption{\label{fig:intro_img}Difference between inverse detection model (top) and CNN-based (bottom) inverse radar models.}
	\end{center}
\end{figure}

The estimation of traversable and occupied areas in an environment is a necessity to reach the goal for cars to behave autonomously. In order to do so, modern cars rely on the combination of different sensors that complement each other. Because of its robustness to environmental changes and its low production price, radars are a major part of current automotive sensor suits \cite{swief2018survey}. To obtain the information of traversable and occupied areas based on the measurements, an inverse sensor model (ISM) has to be applied. The typical way to obtain the ISM would then be to first apply an inverse detection model (IDM) which describes the state of the environment solely based on a single detection and afterwards, accumulate the IDM estimates \cite{thrun2005probabilistic}. Since the summation operation is commutative, the detection's order and thus their spatial correlation gets lost in this process. Additionally, capturing all environment dependent effects for sensors like radars is a non-trivial task. A typical radar IDM and the resulting ISM for the free, occupied and unknown state is shown in the upper part of Fig. \ref{fig:intro_img}.
Because of the uncertainties and the sparseness in the radar data, many measurements have to be accumulated over time to reach certain levels of confidence about the environment. Thus, using sparse radar detections for environment modeling at high velocities for e.g. on highways becomes a difficult task.   

In this work, we address the sparsity problem and the modeling of environment dependent sensor effects of the above explained radar ISM by employing evidential-based convolutional neural networks \cite{sensoy2018evidential}. In order for the model to learn these relationships, we use patches cut out of LiDAR-based occupancy maps according to the vehicles position as targets during training. By doing so, we try to approximate the true state of the environment around the vehicle as close as possible. Moreover, evidential convolutional neural networks are able to decrease the classification probabilities according to the uncertainties in a dataset and shift the reduced amount into a separate unknown state in an end-to-end manner. We further extend this framework to incorporate model uncertainties and show how this improves the robustness of the predictions in contrast to convolutional neural networks without the evidential extension.
%
\section{RELATED WORK}

In this work, we want to describe the occupancy state in a squared area around a vehicle which we will refer to as a scene. A common representation of the scenes states for real-time robotics (\cite{carrillo2015autonomous,fleuret2008multicamera,lombacher2017semantic}) is a discrete 2D bird's-eye-view grid. To describe the occupancy, there are currently two main frameworks, namely the probabilistic framework \cite{elfes1989using} and the Dempster-Shafer theory of evidence (DST) \cite{pagac1998evidential,dempster1968generalization}. Here, the state can either be expressed by the probability of being free/occupied $p_f \in [0,1]$/$p_o=1-p_f$ or alternatively by the belief of being free, occupied or unknown $b_f, b_o, u \in [0,1]$, $b_f+ b_o+ u=1$. By separately modeling the unknown state, the DST is able to express a conflict state additionally to the unknown state. This additional information can aid in tasks like moving object detection, which motivates us to use the DST \cite{moras2011moving}.

\subsection{Inverse Sensor Models based on Inverse Detection Models}
To update a discrete occupancy map given radar measurements, an IDM is typically used. Such an IDM is depicted in Fig. \ref{fig:ism_r} (left) for a noise-free detection of a radial sensor like camera, radar or LiDAR. In the event of uncertainty, several approaches have been proposed. Werber et al. \cite{werber2015automotive} modeled the occupancy state around a detection as Gaussian distribution while all the points in a straight line between the uncertainty ellipse and the sensor are free to a certain extent. This approach, however, results in discontinuous behavior at the boundary between occupied and free space. An alternative was proposed by Loop et al. \cite{loop2016closed}. It describes the incorporation of sensor noise characteristics in a principled way using the Bayesian framework. This is why we propose to use it as a reference for IDMs in this work. A more detailed description is provided in sec. \ref{subsec:ref_hand_radar_ism}.

\subsection{Deep Inverse Sensor Models}
\label{subsec:deep_ism}
Alternatively to using an IDM, the occupation state can be estimated using all detections and their correlations at the same time. Because of the high complexity of such models, they are dominated by deep learning approaches. Training neural networks to learn sensor characteristics has a long history \cite{van1996neural,thrun1993exploration}. However, with the recent progress in stochastic inference in deep learning, models have been proposed to also incorporate uncertainties. Wheeler et al. \cite{wheeler2017deep} addressed training of forward radar models which estimate the radar signal given the state of the environment. To do so, they used a conditional variational Autoencoder (VAE) \cite{sohn2015learning} additionally trained with an adversarial loss \cite{goodfellow2014generative} to reduce blur. Lundell et al. \cite{lundell2018deep} proposed a model to estimate the distance to each untraversable region insight as a Laplace probability density function (PDF). Moreover, they described how to integrate the Laplace distance PDF into an IDM. Recently, Weston et al. \cite{weston2018probably} used a VAE like architecture to infer the state of the environment based on raw, dense radar images. In contrast to the traditional VAE, they modeled the log odds to be Gaussian distributed. To incorporate the uncertainties, they treated the Gaussian log odds as a prior to a Sigmoid and marginalized out the uncertainty.

In this work, we approach the problem from an evidential view and hence need to reformulate the uncertainty estimation and incorporation. However, the majority of proposed deep ISMs use a U-Net architecture with skip connections as a model \cite{ronneberger2015u}, which we also adapt for our work.

\subsection{Uncertainty estimation in Neural Networks}
\label{subsec:prob_dens_est_with_nn}
The uncertainty estimation methods mentioned in Sec. \ref{subsec:deep_ism} model two types of uncertainties, namely the model and data uncertainties which are often referred to as epistemic and aleatoric \cite{kendall2017uncertainties}. Epistemic uncertainties can be estimated in neural networks using sampling techniques e.g. perturbation of weights, inputs or activations \cite{gal2016dropout,goodfellow2014generative,kingma2013auto} or by using ensemble methods \cite{osband2016deep,pearce2018high}. These estimates can e.g. be used to predict how closely a new data point resembles the training set and to assess the robustness of predictions during test time. Aleatoric uncertainties, on the other hand, directly estimate the shape parameters of an a priori defined probability distribution e.g. the mean and variance of a Gaussian \cite{nix1994estimating,lundell2018deep}. One method particularly similar to ours has been proposed by Sensoy et al. \cite{sensoy2018evidential}. They transform an aleatoric, probabilistic classification model into an evidential one using Subjective Logic \cite{josang2016subjective}.

We propose to adapt Sensoy's evidential approach for datasets with explicit unknown regions and extend it to use both aleatoric and epistemic uncertainties.  
%
\section{METHOD}
\label{sec:method}
In this work, we seek to infer the occupancy state of a scene $\vec{s}$ given radar measurements $\vec{x}$. This state can be expressed in a probabilistic \eqref{eq:prob_state} or evidential \eqref{eq:ev_state} frame as follows
\begin{align}	
	\label{eq:prob_state}
	\vec{s}^{(p)} &= \begin{bmatrix} p_f,& p_o \end{bmatrix}^\top=\vec{p}, &&p_f+p_o=1\\
	\label{eq:ev_state}
	\vec{s}^{(e)} &= \begin{bmatrix} b_f,& b_o,& u \end{bmatrix}^\top=\begin{bmatrix} \vec{b},& u \end{bmatrix}^\top, &&b_f+b_o+u=1
\end{align}
with $p_f,p_o$ being the free and occupation probabilities respectively, $u$ the unknown mass and $b_f$, $b_o$ the free and occupation belief masses respectively. We want to compare the IDM-based ISM with two deep learning approaches. 

The first deep learning method consists in treating the problem as a standard three class classification task. Here, we can train a neural network $f_\phi$ to estimate a belief vector which is as close as possible to a target vector $\tilde{\vec{s}}$ given radar inputs. This, however, treats the unknown state as a separate class which neither models nor incorporates the uncertainties of the free and occupied predictions. To ensure comparability, we also apply an epistemic approach to $f_\phi$ and use the expectation $\hat{\vec{y}}$ as an input to the softmax. This architecture is depicted in Fig. \ref{fig:unet_model} (top).

The second method we want to investigate follows Sensoy et al. \cite{sensoy2018evidential}. The method is described for the general case of $K$ classes with an additional unknown class leading to a $K+1$ class classification problem. At the beginning, the unknown state is ignored and a Dirichlet density network \cite{gast2018lightweight,sensoy2018evidential} is used to model the PDF over the remaining $K$ classes. Hence, we model the second order probabilities of our classes which can be used to state how certain the classification probabilities are. More specifically, the Dirichlet distribution's expectation and variance can be interpreted as the classification result and its corresponding aleatoric uncertainty estimate. This is illustrated in the orange boxes on the right in Fig. \ref{fig:unet_model} for two examples which have the same classification probabilities with different uncertainties. 

The remaining question is how to incorporate the additional information about the uncertainty into the classification results. Here, we seek to utilize the uncertainties by reducing each classification probability accordingly and collect all the uncertainty mass in the so far ignored unknown class. This corresponds to a transformation from probabilistic to evidential view which can be described by Subjective Logic \cite{josang2016subjective} in a principled way. 

\subsection{Subjective Logic}
In Subjective Logic, evidence $\vec{e}\in\mathbb{R}_{>0}$ is collected for each of the $K$ classes in the form of strictly positive, real numbers. Given such an evidence vector, Subjective Logic defines the corresponding class beliefs $\vec{b}$ and the unknown state $u$ as follows
\begin{align}
	\label{eq:e2ev}
	\vec{b} &= \dfrac{\vec{e}}{S},\quad u = \dfrac{K}{S},\quad S = K + \sum_{k=1}^{K}e_k
\end{align}

On the probabilistic side, the classification probabilities correspond to the expectation of a Dirichlet PDF whose shape parameters $\vec{\alpha}$ depend on the evidence in the following way
\begin{align}
	\label{eq:e2alpha}
	\vec{\alpha} &= \vec{e} + K\vec{a}\\
	\label{eq:e2prob}
	\vec{p} &= \mathbb{E}[ \text{Dir}(\vec{\alpha}) ] = \dfrac{\vec{e} + K\vec{a}}{S}
\end{align}
Here, the base rate $\vec{a} \in [0,1]$, $\sum_{k=1}^{K}a_k=1$ is the connecting element between the evidential and probabilistic view. It defines how the unknown mass shall be distributed into the $K$ probabilities in the following way
\begin{align}
	\vec{p} &= \vec{b} + u\vec{a}
\end{align} 
We chose $\vec{a} = \dfrac{1}{K}\text{I}$ since we want the probabilities to be equally distributed in the case of complete unknowingness.

We can now use a neural network $f_\theta$ to estimate an evidence vector by e.g. using a quadratic output activation function $\hat{\vec{e}} = f_\theta(\vec{x})^2$. This vector can then be used to define the shape parameters of a Dirichlet distribution \eqref{eq:e2alpha}. The network can then be trained in a way that the Dirichlet's expectation resembles the training data as closely as possible. During test time, the neural network's evidence estimates can be used to compute the belief masses following \eqref{eq:e2ev}.

\subsection{Epistemic Uncertainties}
\begin{figure}
	\begin{center}
		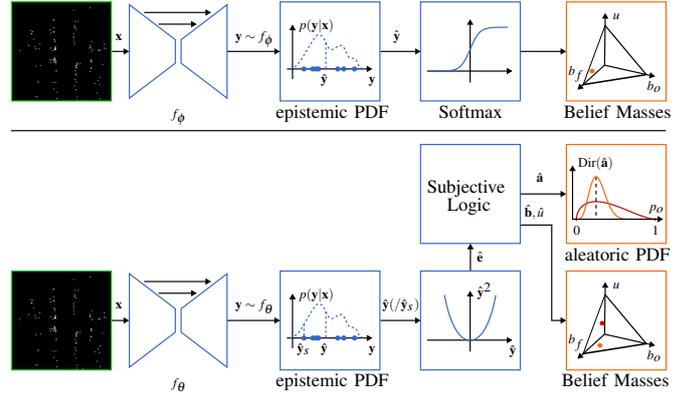
		\caption{\label{fig:unet_model}Comparison of Soft-Net (top) and Ev-Net(-S) (bottom) architectures.}
	\end{center}
\end{figure}

Now we want to describe a method of how to incorporate epistemic uncertainties into this framework. We address this problem by modeling the network's evidence estimates as an epistemic PDF. To do so, any of the methods mentioned in sec. \ref{subsec:prob_dens_est_with_nn} can be applied. The way we want to make the estimations more robust using the epistemic PDF is by taking the $n$th percentile instead of the expectation as the prediction. This converges to the secure unknown state in case of high uncertainty. Moreover, we do not make any assumptions on the PDF's shape. Since epistemic uncertainty prediction methods currently highly rely on sampling, the $n$th percentile can simply be found by sorting the samples (e.g. for $100$ samples the $10$th percentile corresponds to the $10$th smallest sample). In the following, we will refer to the epistemic PDF's expectation as $\hat{\vec{y}}$ and to the $n$th percentile estimate as $\hat{\vec{y}}_{s}$. A complete illustration of the model is depicted in Fig. \ref{fig:unet_model}. 

\subsection{Training the Model}
Finally, we want to address how to train a neural network with a Dirichlet output activation. Several approaches have been proposed e.g. to minimize the Dirichlet's negative log-likelihood \cite{gast2018lightweight} or the Bayesian risk \cite{sensoy2018evidential}. We will use the Bayesian risk with a mean squared loss and Dirichlet prior as shown in \eqref{eq:bayes_risk} since it has been reported to perform with greater stability \cite{gast2018lightweight}.
\begin{align}
	\label{eq:bayes_risk}
	\mathcal{L}(\theta)_i &= \int \|\tilde{\vec{p}}_i  - \hat{\vec{p}}_i\|_{2}^{2} \cdot \text{Dir}(\vec{\alpha}_i(f_\theta(\vec{x}))) d\hat{\vec{p}}_i\\
	&\approx \sum_{k={f,o}} (\tilde{p}_{i,k} - \hat{p}_{i,k})^2 \cdot \dfrac{\hat{p}_{i,k}(1-\hat{p}_{i,k})}{S_i+1}	
\end{align}   
Here $\mathcal{L}(\theta)_i$ is the loss of the $i$th pixel of one training sample, $\tilde{\vec{p}}_i$ are the target class probabilities and $\hat{\vec{p}}_i$ is the expectation of the Dirichlet. This loss ensures that $\hat{\vec{p}}_i$ is as close as possible to $\tilde{\vec{p}}_i$ while at the same time maximizing the accumulated evidence $S_i$ which reduces the uncertainty. Sensoy et al. \cite{sensoy2018evidential} have shown that for this loss the reconstruction part is always favored over the decrease of uncertainty.

\subsection{Unknown Observations}
The problem with the above formulation is that an unknown observation cannot be expressed correctly as a probability. The closest possibility would be to describe the unknown state as $\tilde{\vec{p}} = \begin{bmatrix}0.5&0.5\end{bmatrix}^\top$. However, the network is able to predict areas where the unknown state is caused by occlusion with high certainty. This leads to an increase in evidence being uniformly distributed into both the free and occupied class. Thus, rather than predicting the unknown targets as $\begin{bmatrix}\hat{\vec{b}},&\hat{u}\end{bmatrix}^\top = \begin{bmatrix}0,&0,&1\end{bmatrix}^\top$ the network ends up predicting them to be conflicting areas $\begin{bmatrix}\hat{\vec{b}},&\hat{u}\end{bmatrix}^\top = \begin{bmatrix}0.5,&0.5,&0\end{bmatrix}^\top$. We therefore altered the loss in the following way
\begin{align}
	\mathcal{L}(\theta)_i = \Bigg(\sum_{k={f,o}} &(\tilde{b}_{i,k} - \hat{p}_{i,k})^2 \cdot \dfrac{\hat{p}_{i,k}(1-\hat{p}_{i,k})}{S_i+1}\Bigg) \cdot (\tilde{b}_{i,f}+\tilde{b}_{i,o}) \nonumber\\
		&+ (1-\hat{u}_i)^2 \cdot \tilde{u}_{i}
\end{align}
The first change is that we used the belief masses as targets. Moreover, we introduced a second term that reduces the overall estimated evidence in case of an unknown target by maximizing the unknown state's mass. As targets we allow free $\begin{bmatrix}1,&0,&0\end{bmatrix}^\top$, occupied $\begin{bmatrix}0,&1,&0\end{bmatrix}^\top$, conflict $\begin{bmatrix}0.5,&0.5,&0\end{bmatrix}^\top$ or unknown $\begin{bmatrix}0,&0,&1\end{bmatrix}^\top$. For the first three targets, the loss collapses back to its original version. However, for the fourth case, the first term gets canceled out and the second term is active. This enables the network to utilize unknown observations during training.
%
\section{EXPERIMENTAL SETUP}
\subsection{Data Collection}
To collect the data, a roof-mounted Velodyne HDL-32 LiDAR, four short range, automotive radars each mounted at a corner of the car and the vehicle's dead-reckoning system are used as sensor inputs. The radars are capable of detecting range, angle and relative velocity of up to 64 objects. The driven route resulted in $28843$ training, $4775$ test and $4775$ validation images. During training, the images are additionally augmented by randomly flipping them along the horizontal and vertical axis and rotating them by a random multiple of $90^\circ$.

\subsection{Radar Images}
The radar images are used as inputs for the neural networks. To create them, we used all the detections of the four corner radars falling into a square area around the vehicle. These detections were transformed and discretized into Cartesian pixel coordinates. Moreover, to account for moving objects, the detections with an absolute velocity higher than a certain threshold were identified as dynamic objects and moved into a second layer. Hence, the radar images are of the dimension $128 \times 128 \times 2$ with the first layer containing the static and the second layer the dynamic detections.

\subsection{LiDAR-based Ground Truth Occupancy Patches}
To obtain targets for the training, LiDAR-based occupancy maps were created. In this process, the LiDAR detections were transformed and discretized into Cartesian pixel coordinates and an ideal radial ISM was applied as shown in Fig. \ref{fig:ism_r} (right). As a simplification, only the non-occluded detections were used. The occupancy estimates of each scene were aligned and accumulated using the pose estimates of the vehicle's dead-reckoning system. Afterwards, for each radar image, a patch was cut out of the LiDAR occupancy maps covering the same area. These patches are then used as approximations of the scene's true occupancy beliefs by discretizing the estimates into free, occupied and unknown pixels resulting in images of dimension $128 \times 128 \times 3$. Using the occupancy map patches as opposed to the LiDAR ISM enables the neural network to learn to predict the occupancy in occluded areas based on shape primitives in the data, for example the typical L-shape of partially observed cars.  
%
\section{MODEL}
In this section, we will describe the different approaches to create the ISMs that we will compare in this work.  
\subsection{Reference, Inverse Sensor Model based on an Inverse Detection Model}
\label{subsec:ref_hand_radar_ism}
\begin{figure}
	\begin{center}
		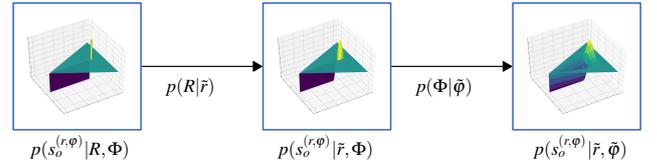
		\caption{\label{fig:ism_r} Visualization of the ideal ISM (left) and the ISMs after integrating the range (mid) and angular noise (right).}
	\end{center}
\end{figure}

The reference we use in this work for an IDM-based radar ISM is based on the work of Loop et al. \cite{loop2016closed} and will be called Ray-ISM. First, we define the condition that an area in the scene at range $r$ and angle $\varphi$ is occupied as follows 
\begin{align}
	\begin{bmatrix}p_f=0,&p_o=1\end{bmatrix}^\top(r,\varphi) =:\vec{s}_o^{(r,\varphi)}
\end{align}
Using the Bayesian framework, the conditional probability of $\vec{s}_o^{(r,\varphi)}$ given range and angular measurements $\tilde{r}, \tilde{\varphi}$ can be expanded to include the true range and angle of the measured object $R, \Phi$ as follows
\begin{align}
	p(s_o^{(r,\varphi)}|\tilde{r},\tilde{\varphi}) &=\iint_{R,\Phi} p(s_o^{(r,\varphi)},R,\Phi|\tilde{r},\tilde{\varphi}) dRd\Phi\\
	&= \iint_{R,\Phi} p(s_o^{(r,\varphi)}|R,\Phi) p(R|\tilde{r}) p(\Phi|\tilde{\varphi}) dRd\Phi
\end{align}
Here, $p(R|\tilde{r}), p(\Phi|\tilde{\varphi})$ are the range and angular noise models respectively. In this work, they are modeled as independent Gaussian noise. Eventually, $p(s_o^{(r,\varphi)}|R,\Phi)$ corresponds to the ideal sensor model which, given the true range and angle, does not depend on the measurements any more. In this work, we extend the ideal ISM from \cite{loop2016closed} to also consider the angle. To do so, we leave the ISM along the range independent of the angle and additionally model the angular component as being equal to the range ISM at $\varphi=\Phi$ and unknown everywhere else as shown in Fig. \ref{fig:ism_r} (left). 
	
Since the range component is independent of the angular, we can first integrate over $R$
\begin{align}
	p(s_o^{(r,\varphi)}|\tilde{r},\tilde{\varphi}) &= \int_{\Phi} p(s_o^{(r,\varphi)}|\Phi,\tilde{r}) p(\Phi|\tilde{\varphi}) d\Phi
\end{align}
and arrive with $p(s_o^{(r,\varphi=\Phi)}|\Phi,\tilde{r}) = p(s_o^{(r)}|\tilde{r})$ at the same model as proposed in \cite{loop2016closed}. The angular integration can then be formulated as 
\begin{align}
	p(s_o^{(r,\varphi)}|&\tilde{r},\tilde{\varphi}) = 0.5\int_{\Phi}\mathcal{N}(\Phi|\tilde{\varphi},\sigma_{\varphi}) d\Phi\\
	 &+\int_{\Phi} (p(s_o^{(r)}|\tilde{r})-0.5)\delta(\varphi-\Phi) \mathcal{N}(\Phi|\tilde{\varphi},\sigma_{\varphi}) d\Phi \nonumber		
\end{align}
Using the equalities
\begin{align}
	\int_{x} \mathcal{N}(x|\mu,\sigma) dx \equiv 1 \text{ and } \int_x \delta(x-a)f(x)dx \equiv f(a)
\end{align}
the final inverse detection model (IDM) for our radar can be written as follows
\begin{align} 
p(s_o^{(r,\varphi)}|\tilde{r},\tilde{\varphi}) &= 0.5+(p(s_o^{(r)}|\tilde{r})-0.5) \mathcal{N}(\varphi|\tilde{\varphi},\sigma_{\varphi})
\end{align}
The ISM can then be obtained by accumulation the IDMs for each detection.

\subsection{Neural Network Models}
\begin{figure}
	\begin{center}
		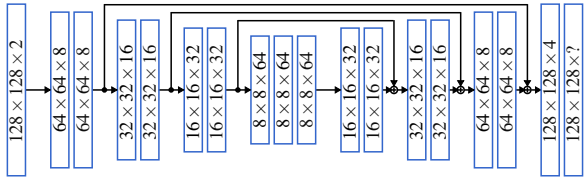
		\caption{\label{fig:nn_architecture} U-Net architecture with convolutions in the encoder and deconvolutions in the decoder. For the Soft-Net, the last layer has three channels while for the Ev-Net/Ev-Net-S it has two.}
	\end{center}
\end{figure}

As mentioned in sec. \ref{subsec:deep_ism}, we are going to follow the work done in the domain of deep ISMs and use an U-Net architecture with skip connections similar to \cite{ronneberger2015u} as shown in Fig. \ref{fig:nn_architecture}. As activations, LeakyReLu are used in all layers expect for the last which is kept linear. All the neural network variants are trained using dropout to make them comparable. We distinguish between three network configurations based on their output activations. 

The first one, which we will refer to as Soft-Net, uses the expectation of the dropout samples $\hat{\vec{y}}$ as an input for a softmax. The softmax activations are used in a cross-entropy loss to train against the belief targets. Hence it has an output dimension of $128 \times 128 \times 3$. 

The second configuration, which we will refer to as Ev-Net, also uses $\hat{\vec{y}}$. However, in this case, the expectations are fed into the evidential framework as shown in Fig. \ref{fig:unet_model} and hence only need two output channels, one for each evidence mass. The last configuration, which we will refer to as Ev-Net-S, is the same as Ev-Net with the difference that instead of using $\hat{\vec{y}}$ it uses $\hat{\vec{y}}_s$.  

%
\section{EXPERIMENTAL RESULTS}
\begin{figure*}
	\begin{center}
		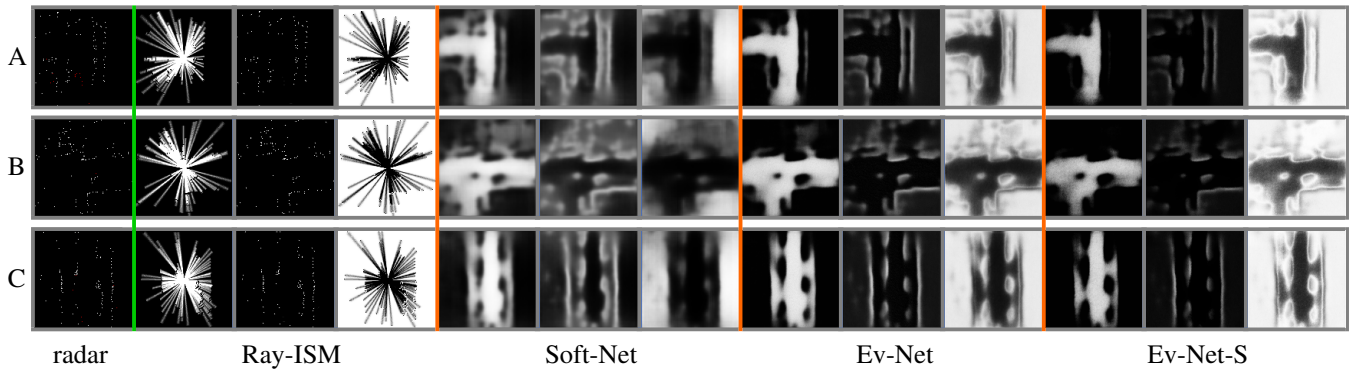
		\caption{\label{fig:results} Results for three different scenes (rows A,B,C) for radar input (1st column) and predictions $\hat{b}_f,\hat{b}_o,\hat{u}$ of the ref ISM (columns 2-4), Ev-Net (columns 5-7), Ev-Net-S (columns 8-10) and Soft-Net (columns 11-13). Here, red detections correspond to dynamic objects (best viewed with zoom on electronic device).}
	\end{center}
\end{figure*}

\subsection{Scores}
To evaluate the model performances we define the conditional probabilities of the estimates given free or occupied targets respectively in \eqref{eq:p_u_j}, \eqref{eq:p_i_j}. Additionally, we define a conflict state and its conditional probability in \eqref{eq:p_c_j}. 
\begin{align}
	\label{eq:p_u_j}
	&p(\hat{u}=\max(\hat{b}_f,\hat{b}_o,\hat{u})|\tilde{b}_j) =: p_{(\hat{u}|j)} \quad\text{, } j=\{\tilde{f},\tilde{o}\}\\
	\label{eq:p_i_j}
	&p(\hat{b}_i=\max(\hat{b}_f,\hat{b}_o,\hat{u})|\tilde{b}_j) =: p_{(i|j)} \quad\text{, } i =\{\hat{f},\hat{o}\}\\
	\label{eq:p_c_j}
	&p(|\hat{b}_{f}-\hat{b}_{o}| \leq 0.2|\tilde{b}_j) =: p_{(\hat{c}|j)}
\end{align}
The conflict is used as a robustness measure of the predictions. Moreover, we separate the scores into visible and hidden targets since their distributions are highly different. The discrimination between hidden and occluded areas is based on the LiDAR measurements. 

\subsection{Quantitative Results}
The scores in Tab. \ref{tab:visible}, \ref{tab:hidden} indicate that Ev-Net-S is too conservative since it reaches the smallest conflict and false rates while at the same time it fails to predict occupied space for both visible and hidden areas. In contrast, Soft-Net performs too aggressively by reaching the highest positive rates while also ending up with the highest conflict and false rates in all neural network approaches. Ev-Net, on the other hand, reaches small false and conflict rates while increasing the performance by a factor of $2.5$ in free and $5$ in occupied space prediction in visible areas relative to the Ray-ISM. In the hidden areas, Ev-Net still outperforms Ray-ISM, even though its performance is reduced. 

\subsection{Qualitative Results}
These observations are also reflected in Fig. \ref{fig:results}. Here, the rows correspond to different scenes, the first column corresponds to the radar input images, the next three columns to the Ray-ISM predictions for $\hat{b}_f, \hat{b}_o, \hat{u}$ followed by the predictions of Soft-Net, Ev-Net and Ev-Net-S. The moving objects are marked as red detections. The chosen scenes are two T-intersections (rows A $\&$ B) and a straight road with parked cars on each side (row C). 

It can be seen that the neural network predictions are much denser than the Ray-ISM's estimates. Moreover, the free and occupied predictions of the Soft-Net are more blurry than the evidential pendants. This is because the Soft-Net's unknown state prediction more closely resembles estimates of hidden areas while the uncertainty about whether an area is occupied or not is distributed into the free and occupied belief masses. In contrast, the evidential networks learn to transfer those uncertainties into the unknown state which results in sharper free/occupied estimates. Moreover, the effect of learned aleatoric uncertainty can clearly be seen at the borders between free and occupied spaces. Here, the sensor noise results in noisy borders causing high unknown estimates. Additionally, the effect of epistemic uncertainty can be seen in areas with complex detection pattern like in the upper left of row B or in the areas behind the parked cars in row C. We also want to mention that moving objects (e.g. red detections at the bottom of the image in row A) do not result in occupied space. 
\begin{table}
	\begin{center}
		\caption{\label{tab:visible}scores for visible area}
		\begin{tabular}{c|C{0.5cm}C{0.5cm}C{0.4cm}|C{0.4cm}||C{0.4cm}C{0.4cm}C{0.4cm}|C{0.4cm}}
			$[\%]$ & $p_{(\hat{f}|\tilde{f})}$ & $p_{(\hat{o}|\tilde{f})}$ & $p_{(\hat{u}|\tilde{f})}$ & $p_{(\hat{c}|\tilde{f})}$  
			& $p_{(\hat{f}|\tilde{o})}$ & $p_{(\hat{o}|\tilde{o})}$ & $p_{(\hat{u}|\tilde{o})}$ & $p_{(\hat{c}|\tilde{o})}$\\ 
			\hline 
			 Ray-ISM &  25 &  0 &  75 &  0 &  25 &  4 &  71 &  0\\
			 Soft-Net &  76 &  16 &  8 &  19 &  17 &  71 &  12 &  19\\
			 Ev-Net  &  64 &  3 &  33 &  1 &  8 &  20 &  72 &  0\\
 			 Ev-Net-S &  48 &  0 &  52 &  0 &  3 &  2 &  95 &  0
		\end{tabular}
		
	\end{center}
	\begin{center}
		\caption{\label{tab:hidden}scores for hidden area}
		\begin{tabular}{c|C{0.4cm}C{0.4cm}C{0.4cm}|C{0.4cm}||C{0.4cm}C{0.4cm}C{0.4cm}|C{0.4cm}}
			$[\%]$ & $p_{(\hat{f}|\tilde{f})}$ & $p_{(\hat{o}|\tilde{f})}$ & $p_{(\hat{u}|\tilde{f})}$ & $p_{(\hat{c}|\tilde{f})}$  
			& $p_{(\hat{f}|\tilde{o})}$ & $p_{(\hat{o}|\tilde{o})}$ & $p_{(\hat{u}|\tilde{o})}$ & $p_{(\hat{c}|\tilde{o})}$\\ 
			\hline 
			Ray-ISM & 9 & 0 & 91 & 0 & 12 & 1 & 87 & 0\\
			Soft-Net & 32 & 32 & 36 & 45 & 14 & 53 & 33 & 35\\			
			Ev-Net & 10 & 1 & 89 & 0 & 3 & 4 & 93 & 0\\
			Ev-Net-S & 3 & 0 & 97 & 0 & 0 & 0 & 100 & 0
		\end{tabular}
		
	\end{center}	
\end{table}  

%
\section{CONCLUSION}
In this work, we addressed the problem of describing the occupation state of an environment based on automotive radar measurements which provide sparse, noisy data. This problem is approached by employing the recently proposed evidential convolutional neural networks which can be used to infer a probability density function (PDF) over the classification result rather than a point prediction. Moreover, the uncertainty represented in the PDF can be used to reduce the classification probabilities accordingly by moving the mass to a separate unknown state. We have shown that this formulation provides an alternative to the typically applied detection-based inverse radar models by improving the performance by a factor of $2.5$ in detecting free and a factor of $5$ in detecting occupied space on our data set. At the same time, in contrast to typical softmax-based classification networks, the model remains robust to uncertainties and reaches low false and conflict rates. Moreover, we have shown how to utilize more spatial information by using occupancy map patches instead of the LiDAR-ISM as targets during training.

%
%
%
%
\bibliographystyle{IEEEtran}
\bibliography{IEEEabrv,bib}
\end{document}

%% file: intro_img.pdf_tex
\begingroup%
  \makeatletter%
  \providecommand\color[2][]{%
    \errmessage{(Inkscape) Color is used for the text in Inkscape, but the package 'color.sty' is not loaded}%
    \renewcommand\color[2][]{}%
  }%
  \providecommand\transparent[1]{%
    \errmessage{(Inkscape) Transparency is used (non-zero) for the text in Inkscape, but the package 'transparent.sty' is not loaded}%
    \renewcommand\transparent[1]{}%
  }%
  \providecommand\rotatebox[2]{#2}%
  \ifx\svgwidth\undefined%
    \setlength{\unitlength}{234.24564392bp}%
    \ifx\svgscale\undefined%
      \relax%
    \else%
      \setlength{\unitlength}{\unitlength * \real{\svgscale}}%
    \fi%
  \else%
    \setlength{\unitlength}{\svgwidth}%
  \fi%
  \global\let\svgwidth\undefined%
  \global\let\svgscale\undefined%
  \makeatother%
  \begin{picture}(1,0.34523409)%
    \put(0,0){\includegraphics[width=\unitlength,page=1]{intro_img.pdf}}%
  \end{picture}%
\endgroup%

%% file: unet_model_softmax.pdf_tex
\begingroup%
  \makeatletter%
  \providecommand\color[2][]{%
    \errmessage{(Inkscape) Color is used for the text in Inkscape, but the package 'color.sty' is not loaded}%
    \renewcommand\color[2][]{}%
  }%
  \providecommand\transparent[1]{%
    \errmessage{(Inkscape) Transparency is used (non-zero) for the text in Inkscape, but the package 'transparent.sty' is not loaded}%
    \renewcommand\transparent[1]{}%
  }%
  \providecommand\rotatebox[2]{#2}%
  \newcommand*\fsize{\dimexpr\f@size pt\relax}%
  \newcommand*\lineheight[1]{\fontsize{\fsize}{#1\fsize}\selectfont}%
  \ifx\svgwidth\undefined%
    \setlength{\unitlength}{244.56163301bp}%
    \ifx\svgscale\undefined%
      \relax%
    \else%
      \setlength{\unitlength}{\unitlength * \real{\svgscale}}%
    \fi%
  \else%
    \setlength{\unitlength}{\svgwidth}%
  \fi%
  \global\let\svgwidth\undefined%
  \global\let\svgscale\undefined%
  \makeatother%
  \begin{picture}(1,0.57095532)%
    \lineheight{1}%
    \setlength\tabcolsep{0pt}%
    \put(0.15945744,0.09843891){\color[rgb]{0,0,0}\makebox(0,0)[lt]{\lineheight{1.25}\smash{\begin{tabular}[t]{l}\tiny $\vec{x}$\end{tabular}}}}%
    \put(0.24417893,-0.0212434){\color[rgb]{0,0,0}\makebox(0,0)[lt]{\lineheight{1.25}\smash{\begin{tabular}[t]{l}\tiny $f_\theta$\end{tabular}}}}%
    \put(0,0){\includegraphics[width=\unitlength,page=1]{unet_model_softmax.pdf}}%
    \put(0.15945744,0.51189939){\color[rgb]{0,0,0}\makebox(0,0)[lt]{\lineheight{1.25}\smash{\begin{tabular}[t]{l}\tiny $\vec{x}$\end{tabular}}}}%
    \put(0,0){\includegraphics[width=\unitlength,page=2]{unet_model_softmax.pdf}}%
    \put(0.24417893,0.39229698){\color[rgb]{0,0,0}\makebox(0,0)[lt]{\lineheight{1.25}\smash{\begin{tabular}[t]{l}\tiny $f_\phi$\end{tabular}}}}%
    \put(0.40429731,-0.02108367){\color[rgb]{0,0,0}\makebox(0,0)[lt]{\lineheight{1.25}\smash{\begin{tabular}[t]{l}\scriptsize epistemic PDF\end{tabular}}}}%
    \put(0.84463712,0.17337976){\color[rgb]{0,0,0}\makebox(0,0)[lt]{\lineheight{1.25}\smash{\begin{tabular}[t]{l}\scriptsize aleatoric PDF\end{tabular}}}}%
    \put(0,0){\includegraphics[width=\unitlength,page=3]{unet_model_softmax.pdf}}%
    \put(0.56660743,0.09843891){\color[rgb]{0,0,0}\makebox(0,0)[lt]{\lineheight{1.25}\smash{\begin{tabular}[t]{l}\tiny $\hat{\vec{y}}(/\hat{\vec{y}}_{s})$\end{tabular}}}}%
    \put(0,0){\includegraphics[width=\unitlength,page=4]{unet_model_softmax.pdf}}%
    \put(0.76002229,0.02803947){\color[rgb]{0,0,0}\makebox(0,0)[lt]{\lineheight{1.25}\smash{\begin{tabular}[t]{l}\tiny$\hat{\vec{y}}$\end{tabular}}}}%
    \put(0,0){\includegraphics[width=\unitlength,page=5]{unet_model_softmax.pdf}}%
    \put(0.71099279,0.11793109){\color[rgb]{0,0,0}\makebox(0,0)[lt]{\lineheight{1.25}\smash{\begin{tabular}[t]{l}\tiny$\hat{\vec{y}}^2$\end{tabular}}}}%
    \put(0.71087913,0.17187733){\color[rgb]{0,0,0}\makebox(0,0)[lt]{\lineheight{1.25}\smash{\begin{tabular}[t]{l}\tiny $\hat{\vec{e}}$\end{tabular}}}}%
    \put(0.97026187,0.02119304){\color[rgb]{0,0,0}\makebox(0,0)[lt]{\lineheight{1.25}\smash{\begin{tabular}[t]{l}\tiny$b_o$\end{tabular}}}}%
    \put(0.85020397,0.0450009){\color[rgb]{0,0,0}\makebox(0,0)[lt]{\lineheight{1.25}\smash{\begin{tabular}[t]{l}\tiny$b_f$\end{tabular}}}}%
    \put(0.91826066,0.12948345){\color[rgb]{0,0,0}\makebox(0,0)[lt]{\lineheight{1.25}\smash{\begin{tabular}[t]{l}\tiny$u$\end{tabular}}}}%
    \put(0,0){\includegraphics[width=\unitlength,page=6]{unet_model_softmax.pdf}}%
    \put(0.97292297,0.24912314){\color[rgb]{0,0,0}\makebox(0,0)[lt]{\lineheight{1.25}\smash{\begin{tabular}[t]{l}\tiny$p_o$\end{tabular}}}}%
    \put(0.97779053,0.21237586){\color[rgb]{0,0,0}\makebox(0,0)[lt]{\lineheight{1.25}\smash{\begin{tabular}[t]{l}\tiny 1\end{tabular}}}}%
    \put(0.8580366,0.21237586){\color[rgb]{0,0,0}\makebox(0,0)[lt]{\lineheight{1.25}\smash{\begin{tabular}[t]{l}\tiny 0\end{tabular}}}}%
    \put(0,0){\includegraphics[width=\unitlength,page=7]{unet_model_softmax.pdf}}%
    \put(0.84006683,-0.02116355){\color[rgb]{0,0,0}\makebox(0,0)[lt]{\lineheight{1.25}\smash{\begin{tabular}[t]{l}\scriptsize Belief Masses\end{tabular}}}}%
    \put(0.652701,0.39237685){\color[rgb]{0,0,0}\makebox(0,0)[lt]{\lineheight{1.25}\smash{\begin{tabular}[t]{l}\scriptsize Softmax\end{tabular}}}}%
    \put(0.97026182,0.43465352){\color[rgb]{0,0,0}\makebox(0,0)[lt]{\lineheight{1.25}\smash{\begin{tabular}[t]{l}\tiny$b_o$\end{tabular}}}}%
    \put(0.85020392,0.45846137){\color[rgb]{0,0,0}\makebox(0,0)[lt]{\lineheight{1.25}\smash{\begin{tabular}[t]{l}\tiny$b_f$\end{tabular}}}}%
    \put(0.91826062,0.54294392){\color[rgb]{0,0,0}\makebox(0,0)[lt]{\lineheight{1.25}\smash{\begin{tabular}[t]{l}\tiny$u$\end{tabular}}}}%
    \put(0,0){\includegraphics[width=\unitlength,page=8]{unet_model_softmax.pdf}}%
    \put(0.58347434,0.51189937){\color[rgb]{0,0,0}\makebox(0,0)[lt]{\lineheight{1.25}\smash{\begin{tabular}[t]{l}\tiny $\hat{\vec{y}}$\end{tabular}}}}%
    \put(0,0){\includegraphics[width=\unitlength,page=9]{unet_model_softmax.pdf}}%
    \put(0.40429731,0.39237685){\color[rgb]{0,0,0}\makebox(0,0)[lt]{\lineheight{1.25}\smash{\begin{tabular}[t]{l}\scriptsize epistemic PDF\end{tabular}}}}%
    \put(0.84310414,0.39229699){\color[rgb]{0,0,0}\makebox(0,0)[lt]{\lineheight{1.25}\smash{\begin{tabular}[t]{l}\scriptsize Belief Masses\end{tabular}}}}%
    \put(0,0){\includegraphics[width=\unitlength,page=10]{unet_model_softmax.pdf}}%
    \put(0.63433005,0.27662491){\color[rgb]{0,0,0}\makebox(0,0)[lt]{\lineheight{1.25}\smash{\begin{tabular}[t]{l}\scriptsize Subjective\end{tabular}}}}%
    \put(0.66499717,0.24855332){\color[rgb]{0,0,0}\makebox(0,0)[lt]{\lineheight{1.25}\smash{\begin{tabular}[t]{l}\scriptsize Logic\end{tabular}}}}%
    \put(0,0){\includegraphics[width=\unitlength,page=11]{unet_model_softmax.pdf}}%
    \put(0.3424097,0.09843891){\color[rgb]{0,0,0}\makebox(0,0)[lt]{\lineheight{1.25}\smash{\begin{tabular}[t]{l}\tiny $\vec{y} \sim f_\theta$\end{tabular}}}}%
    \put(0,0){\includegraphics[width=\unitlength,page=12]{unet_model_softmax.pdf}}%
    \put(0.34240971,0.51189937){\color[rgb]{0,0,0}\makebox(0,0)[lt]{\lineheight{1.25}\smash{\begin{tabular}[t]{l}\tiny $\vec{y} \sim f_\phi$\end{tabular}}}}%
    \put(0,0){\includegraphics[width=\unitlength,page=13]{unet_model_softmax.pdf}}%
    \put(0.54543895,0.44483483){\color[rgb]{0,0,0}\makebox(0,0)[lt]{\lineheight{1.25}\smash{\begin{tabular}[t]{l}\tiny $\vec{y}$\end{tabular}}}}%
    \put(0.43875315,0.5276361){\color[rgb]{0,0,0}\makebox(0,0)[lt]{\lineheight{1.25}\smash{\begin{tabular}[t]{l}\tiny $p(\vec{y}|\vec{x})$\end{tabular}}}}%
    \put(0,0){\includegraphics[width=\unitlength,page=14]{unet_model_softmax.pdf}}%
    \put(0.47273304,0.44483483){\color[rgb]{0,0,0}\makebox(0,0)[lt]{\lineheight{1.25}\smash{\begin{tabular}[t]{l}\tiny $\hat{\vec{y}}$\end{tabular}}}}%
    \put(0,0){\includegraphics[width=\unitlength,page=15]{unet_model_softmax.pdf}}%
    \put(0.54543895,0.03137433){\color[rgb]{0,0,0}\makebox(0,0)[lt]{\lineheight{1.25}\smash{\begin{tabular}[t]{l}\tiny $\vec{y}$\end{tabular}}}}%
    \put(0.43875315,0.1141756){\color[rgb]{0,0,0}\makebox(0,0)[lt]{\lineheight{1.25}\smash{\begin{tabular}[t]{l}\tiny $p(\vec{y}|\vec{x})$\end{tabular}}}}%
    \put(0,0){\includegraphics[width=\unitlength,page=16]{unet_model_softmax.pdf}}%
    \put(0.47273304,0.03137433){\color[rgb]{0,0,0}\makebox(0,0)[lt]{\lineheight{1.25}\smash{\begin{tabular}[t]{l}\tiny $\hat{\vec{y}}$\end{tabular}}}}%
    \put(0,0){\includegraphics[width=\unitlength,page=17]{unet_model_softmax.pdf}}%
    \put(0.43618042,0.03137433){\color[rgb]{0,0,0}\makebox(0,0)[lt]{\lineheight{1.25}\smash{\begin{tabular}[t]{l}\tiny $\hat{\vec{y}}_s$\end{tabular}}}}%
    \put(0.80177492,0.29337471){\color[rgb]{0,0,0}\makebox(0,0)[lt]{\lineheight{1.25}\smash{\begin{tabular}[t]{l}\tiny $\hat{\vec{a}}$\end{tabular}}}}%
    \put(0.86502581,0.31599171){\color[rgb]{0,0,0}\makebox(0,0)[lt]{\lineheight{1.25}\smash{\begin{tabular}[t]{l}\tiny $\text{Dir}(\hat{\vec{a}})$\end{tabular}}}}%
    \put(0.78337464,0.2394632){\color[rgb]{0,0,0}\makebox(0,0)[lt]{\lineheight{1.25}\smash{\begin{tabular}[t]{l}\tiny $\hat{\vec{b}}, \hat{u}$\end{tabular}}}}%
    \put(0,0){\includegraphics[width=\unitlength,page=18]{unet_model_softmax.pdf}}%
  \end{picture}%
\endgroup%

%% file: ism.pdf_tex
\begingroup%
  \makeatletter%
  \providecommand\color[2][]{%
    \errmessage{(Inkscape) Color is used for the text in Inkscape, but the package 'color.sty' is not loaded}%
    \renewcommand\color[2][]{}%
  }%
  \providecommand\transparent[1]{%
    \errmessage{(Inkscape) Transparency is used (non-zero) for the text in Inkscape, but the package 'transparent.sty' is not loaded}%
    \renewcommand\transparent[1]{}%
  }%
  \providecommand\rotatebox[2]{#2}%
  \ifx\svgwidth\undefined%
    \setlength{\unitlength}{237.56833775bp}%
    \ifx\svgscale\undefined%
      \relax%
    \else%
      \setlength{\unitlength}{\unitlength * \real{\svgscale}}%
    \fi%
  \else%
    \setlength{\unitlength}{\svgwidth}%
  \fi%
  \global\let\svgwidth\undefined%
  \global\let\svgscale\undefined%
  \makeatother%
  \begin{picture}(1,0.2446043)%
    \put(0.42292833,0.00210765){\color[rgb]{0,0,0}\makebox(0,0)[lb]{\smash{\scriptsize $p(s_o^{(r,\varphi)}|\tilde{r},\Phi)$}}}%
    \put(0.82414943,0.00210765){\color[rgb]{0,0,0}\makebox(0,0)[lb]{\smash{\scriptsize $p(s_o^{(r,\varphi)}|\tilde{r},\tilde{\varphi})$}}}%
    \put(0.24660442,0.1014048){\color[rgb]{0,0,0}\makebox(0,0)[lb]{\smash{\scriptsize $p(R|\tilde{r})$}}}%
    \put(0.63872968,0.1014048){\color[rgb]{0,0,0}\makebox(0,0)[lb]{\smash{\scriptsize $p(\Phi|\tilde{\varphi})$}}}%
    \put(0.03433524,0.00210765){\color[rgb]{0,0,0}\makebox(0,0)[lb]{\smash{\scriptsize $p(s_o^{(r,\varphi)}|R,\Phi)$}}}%
    \put(0,0){\includegraphics[width=\unitlength,page=1]{ism.pdf}}%
  \end{picture}%
\endgroup%

%% file: nn_architecture.pdf_tex
\begingroup%
  \makeatletter%
  \providecommand\color[2][]{%
    \errmessage{(Inkscape) Color is used for the text in Inkscape, but the package 'color.sty' is not loaded}%
    \renewcommand\color[2][]{}%
  }%
  \providecommand\transparent[1]{%
    \errmessage{(Inkscape) Transparency is used (non-zero) for the text in Inkscape, but the package 'transparent.sty' is not loaded}%
    \renewcommand\transparent[1]{}%
  }%
  \providecommand\rotatebox[2]{#2}%
  \newcommand*\fsize{\dimexpr\f@size pt\relax}%
  \newcommand*\lineheight[1]{\fontsize{\fsize}{#1\fsize}\selectfont}%
  \ifx\svgwidth\undefined%
    \setlength{\unitlength}{215.08444959bp}%
    \ifx\svgscale\undefined%
      \relax%
    \else%
      \setlength{\unitlength}{\unitlength * \real{\svgscale}}%
    \fi%
  \else%
    \setlength{\unitlength}{\svgwidth}%
  \fi%
  \global\let\svgwidth\undefined%
  \global\let\svgscale\undefined%
  \makeatother%
  \begin{picture}(1,0.30251813)%
    \lineheight{1}%
    \setlength\tabcolsep{0pt}%
    \put(0.02688416,0.05774125){\color[rgb]{0,0,0}\rotatebox{90}{\makebox(0,0)[lt]{\lineheight{1.25}\smash{\begin{tabular}[t]{l}\scriptsize $128 \times 128 \times 2$\end{tabular}}}}}%
    \put(0,0){\includegraphics[width=\unitlength,page=1]{nn_architecture.pdf}}%
    \put(0.10245403,0.07782146){\color[rgb]{0,0,0}\rotatebox{90}{\makebox(0,0)[lt]{\lineheight{1.25}\smash{\begin{tabular}[t]{l}\scriptsize $64 \times 64 \times 8$\end{tabular}}}}}%
    \put(0.14286703,0.07782146){\color[rgb]{0,0,0}\rotatebox{90}{\makebox(0,0)[lt]{\lineheight{1.25}\smash{\begin{tabular}[t]{l}\scriptsize $64 \times 64 \times 8$\end{tabular}}}}}%
    \put(0,0){\includegraphics[width=\unitlength,page=2]{nn_architecture.pdf}}%
    \put(0.21815009,0.06387344){\color[rgb]{0,0,0}\rotatebox{90}{\makebox(0,0)[lt]{\lineheight{1.25}\smash{\begin{tabular}[t]{l}\scriptsize $32 \times 32 \times 16$\end{tabular}}}}}%
    \put(0.25856305,0.06387344){\color[rgb]{0,0,0}\rotatebox{90}{\makebox(0,0)[lt]{\lineheight{1.25}\smash{\begin{tabular}[t]{l}\scriptsize $32 \times 32 \times 16$\end{tabular}}}}}%
    \put(0,0){\includegraphics[width=\unitlength,page=3]{nn_architecture.pdf}}%
    \put(0.33384604,0.06387344){\color[rgb]{0,0,0}\rotatebox{90}{\makebox(0,0)[lt]{\lineheight{1.25}\smash{\begin{tabular}[t]{l}\scriptsize $16 \times 16 \times 32$\end{tabular}}}}}%
    \put(0.374259,0.06387344){\color[rgb]{0,0,0}\rotatebox{90}{\makebox(0,0)[lt]{\lineheight{1.25}\smash{\begin{tabular}[t]{l}\scriptsize $16 \times 16 \times 32$\end{tabular}}}}}%
    \put(0,0){\includegraphics[width=\unitlength,page=4]{nn_architecture.pdf}}%
    \put(0.44954197,0.08479543){\color[rgb]{0,0,0}\rotatebox{90}{\makebox(0,0)[lt]{\lineheight{1.25}\smash{\begin{tabular}[t]{l}\scriptsize $8 \times 8 \times 64$\end{tabular}}}}}%
    \put(0,0){\includegraphics[width=\unitlength,page=5]{nn_architecture.pdf}}%
    \put(0.48995491,0.08479543){\color[rgb]{0,0,0}\rotatebox{90}{\makebox(0,0)[lt]{\lineheight{1.25}\smash{\begin{tabular}[t]{l}\scriptsize $8 \times 8 \times 64$\end{tabular}}}}}%
    \put(0.53036795,0.08479543){\color[rgb]{0,0,0}\rotatebox{90}{\makebox(0,0)[lt]{\lineheight{1.25}\smash{\begin{tabular}[t]{l}\scriptsize $8 \times 8 \times 64$\end{tabular}}}}}%
    \put(0,0){\includegraphics[width=\unitlength,page=6]{nn_architecture.pdf}}%
    \put(0.60565108,0.06387344){\color[rgb]{0,0,0}\rotatebox{90}{\makebox(0,0)[lt]{\lineheight{1.25}\smash{\begin{tabular}[t]{l}\scriptsize $16 \times 16 \times 32$\end{tabular}}}}}%
    \put(0.64606403,0.06387344){\color[rgb]{0,0,0}\rotatebox{90}{\makebox(0,0)[lt]{\lineheight{1.25}\smash{\begin{tabular}[t]{l}\scriptsize $16 \times 16 \times 32$\end{tabular}}}}}%
    \put(0,0){\includegraphics[width=\unitlength,page=7]{nn_architecture.pdf}}%
    \put(0.72134706,0.06387344){\color[rgb]{0,0,0}\rotatebox{90}{\makebox(0,0)[lt]{\lineheight{1.25}\smash{\begin{tabular}[t]{l}\scriptsize $32 \times 32 \times 16$\end{tabular}}}}}%
    \put(0.76176006,0.06387344){\color[rgb]{0,0,0}\rotatebox{90}{\makebox(0,0)[lt]{\lineheight{1.25}\smash{\begin{tabular}[t]{l}\scriptsize $32 \times 32 \times 16$\end{tabular}}}}}%
    \put(0,0){\includegraphics[width=\unitlength,page=8]{nn_architecture.pdf}}%
    \put(0.83704309,0.07782146){\color[rgb]{0,0,0}\rotatebox{90}{\makebox(0,0)[lt]{\lineheight{1.25}\smash{\begin{tabular}[t]{l}\scriptsize $64 \times 64 \times 8$\end{tabular}}}}}%
    \put(0.87745609,0.07782146){\color[rgb]{0,0,0}\rotatebox{90}{\makebox(0,0)[lt]{\lineheight{1.25}\smash{\begin{tabular}[t]{l}\scriptsize $64 \times 64 \times 8$\end{tabular}}}}}%
    \put(0.95245228,0.05774126){\color[rgb]{0,0,0}\rotatebox{90}{\makebox(0,0)[lt]{\lineheight{1.25}\smash{\begin{tabular}[t]{l}\scriptsize $128 \times 128 \times 4$\end{tabular}}}}}%
    \put(0,0){\includegraphics[width=\unitlength,page=9]{nn_architecture.pdf}}%
    \put(0.99286528,0.05774126){\color[rgb]{0,0,0}\rotatebox{90}{\makebox(0,0)[lt]{\lineheight{1.25}\smash{\begin{tabular}[t]{l}\scriptsize $128 \times 128 \times ?$\end{tabular}}}}}%
    \put(0,0){\includegraphics[width=\unitlength,page=10]{nn_architecture.pdf}}%
  \end{picture}%
\endgroup%

%% file: results.pdf_tex
\begingroup%
  \makeatletter%
  \providecommand\color[2][]{%
    \errmessage{(Inkscape) Color is used for the text in Inkscape, but the package 'color.sty' is not loaded}%
    \renewcommand\color[2][]{}%
  }%
  \providecommand\transparent[1]{%
    \errmessage{(Inkscape) Transparency is used (non-zero) for the text in Inkscape, but the package 'transparent.sty' is not loaded}%
    \renewcommand\transparent[1]{}%
  }%
  \providecommand\rotatebox[2]{#2}%
  \newcommand*\fsize{\dimexpr\f@size pt\relax}%
  \newcommand*\lineheight[1]{\fontsize{\fsize}{#1\fsize}\selectfont}%
  \ifx\svgwidth\undefined%
    \setlength{\unitlength}{497.16037059bp}%
    \ifx\svgscale\undefined%
      \relax%
    \else%
      \setlength{\unitlength}{\unitlength * \real{\svgscale}}%
    \fi%
  \else%
    \setlength{\unitlength}{\svgwidth}%
  \fi%
  \global\let\svgwidth\undefined%
  \global\let\svgscale\undefined%
  \makeatother%
  \begin{picture}(1,0.28249891)%
    \lineheight{1}%
    \setlength\tabcolsep{0pt}%
    \put(0,0){\includegraphics[width=\unitlength,page=1]{results.pdf}}%
    \put(0.63050723,0.01187261){\color[rgb]{0,0,0}\makebox(0,0)[lt]{\lineheight{1.25}\smash{\begin{tabular}[t]{l}Ev-Net\end{tabular}}}}%
    \put(0.84895248,0.01187261){\color[rgb]{0,0,0}\makebox(0,0)[lt]{\lineheight{1.25}\smash{\begin{tabular}[t]{l}Ev-Net-S\end{tabular}}}}%
    \put(0,0){\includegraphics[width=\unitlength,page=2]{results.pdf}}%
    \put(0.16947952,0.01187261){\color[rgb]{0,0,0}\makebox(0,0)[lt]{\lineheight{1.25}\smash{\begin{tabular}[t]{l}Ray-ISM\end{tabular}}}}%
    \put(0.02796837,0.01187261){\color[rgb]{0,0,0}\makebox(0,0)[lt]{\lineheight{1.25}\smash{\begin{tabular}[t]{l}radar\end{tabular}}}}%
    \put(-0.00613642,0.23717405){\color[rgb]{0,0,0}\makebox(0,0)[lt]{\lineheight{1.25}\smash{\begin{tabular}[t]{l}A\end{tabular}}}}%
    \put(-0.00613642,0.15391196){\color[rgb]{0,0,0}\makebox(0,0)[lt]{\lineheight{1.25}\smash{\begin{tabular}[t]{l}B\end{tabular}}}}%
    \put(-0.00613642,0.06952718){\color[rgb]{0,0,0}\makebox(0,0)[lt]{\lineheight{1.25}\smash{\begin{tabular}[t]{l}C\end{tabular}}}}%
    \put(0,0){\includegraphics[width=\unitlength,page=3]{results.pdf}}%
    \put(0.39699298,0.01189328){\color[rgb]{0,0,0}\makebox(0,0)[lt]{\lineheight{1.25}\smash{\begin{tabular}[t]{l}Soft-Net\end{tabular}}}}%
    \put(0,0){\includegraphics[width=\unitlength,page=4]{results.pdf}}%
  \end{picture}%
\endgroup%